\documentclass[10pt,twocolumn,letterpaper]{article}

\usepackage{cvpr}              
\usepackage{marvosym}
\definecolor{cvprblue}{rgb}{0.21,0.49,0.74}
\usepackage[pagebackref,breaklinks,colorlinks,allcolors=cvprblue]{hyperref}

\def\paperID{8} 
\def\confName{CVPR}
\def\confYear{2026}

\title{Quantization with Unified Adaptive Distillation to enable multi-LoRA based one-for-all Generative Vision Models on edge}

\author{Sowmya Vajrala$^{*,1, \textsuperscript{\Letter}}$, 
Aakash Parmar$^{*,1}$, 
Prasanna R$^{*,1}$, 
Sravanth Kodavanti$^{1}$, 
Manjunath Arveti$^{1}$, \\
Srinivas Soumitri Miriyala$^{1}$, 
Ashok Senapati$^{1}$ \\
$^{1}$Samsung Research Institute Bangalore, India \\
}

\begin{document}
\maketitle

\def\thefootnote{}\footnotetext{
*Equal Contribution
\quad\textsuperscript{\Letter}v.lahari@samsung.com \\
}
\addtocounter{footnote}{-1} 


\begin{abstract}
Generative Artificial Intelligence (GenAI) features such as image editing, object removal, and prompt-guided image transformation are increasingly integrated into mobile applications. However, deploying Large Vision Models (LVMs) for such tasks on resource-constrained devices remains challenging due to their high memory and compute requirements. While Low-Rank Adapters (LoRAs) enable parameter-efficient task adaptation, existing Mobile deployment pipelines typically compile separate model binaries for each LoRA + a copy of the foundation model, resulting in redundant storage and increased runtime overhead. In this work, we present a unified framework for enabling multi-task GenAI inference on edge devices using a single shared model. Our key idea is to treat LoRA weights as runtime inputs rather than embedding them into the compiled model graph, allowing dynamic task switching at runtime without recompilation. Then, to support efficient on-device execution, we introduce QUAD (Quantization with Unified Adaptive Distillation), a quantization-aware training strategy that aligns multiple LoRA adapters under a shared quantization profile. We implement the proposed system with a lightweight runtime stack compatible with mobile NPUs and evaluate it across multiple chipsets. Experimental results demonstrate up to $6\times$ and $4\times$ reduction in memory footprint and latency improvements, respectively, while maintaining high visual quality across multiple GenAI tasks.
\end{abstract}    
\section{Introduction}
\label{sec:intro}


Generative Artificial Intelligence (GenAI) is rapidly transforming image creation and editing on consumer devices. Modern applications now enable users to remove unwanted objects, enhance photo quality, modify backgrounds, or generate images from textual prompts. These capabilities are powered by Large Vision Models (LVMs) trained on large-scale visual data that can perform tasks such as inpainting ~\cite{quan2024deep}, prompt-guided stylization ~\cite{liu2023portrait}, and image-to-image translation ~\cite{xia2024diffi2i}. 
Despite their effectiveness, LVMs are computationally expensive and memory-intensive, which limits their deployment on resource-constrained devices such as smartphones. As a result, most GenAI services today rely on cloud-based inference. In such systems, user inputs are transmitted to remote servers where the model performs inference and returns the generated result. However, this approach introduces latency due to network communication, raises privacy concerns related to user data transmission, and requires a reliable internet connection. These limitations have motivated growing interest in on-device GenAI, where models execute directly on edge hardware.

Deploying LVMs on mobile devices presents several challenges due to limited compute, memory, and storage resources. Moreover, many applications require task-specific adaptations of a shared base model. A widely adopted solution is the use of Low-Rank Adapters (LoRAs) ~\cite{mao2025survey}, which enable parameter-efficient fine-tuning by learning low-rank updates while keeping the base model frozen. This allows multiple task-specific adapters to be trained with minimal additional parameters ~\cite{xin2024parameter}.

\begin{figure}
    \centering
    \includegraphics[width=\linewidth]{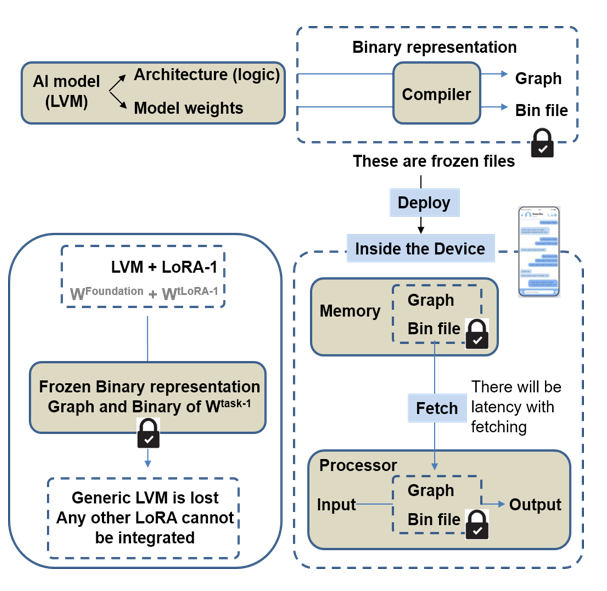}
    \caption{Schematic for a simplified version of compilation and deployment of LVM and depicting the problem definition}
    \label{fig:fig1}
\end{figure}

While the LoRA paradigm enables rapid specialization, it introduces significant deployment challenges on edge hardware. In particular, when each LoRA adapter is quantized independently, the resulting adapters often require different quantization parameters (e.g., scale and zero-point), making them incompatible with a single static inference graph. This prevents efficient runtime switching between tasks, increases memory overhead due to multiple calibration states, and complicates deployment on NPUs where fixed quantization parameters are typically required (see Figure. ~\ref{fig:fig1}).

Given a pretrained diffusion backbone and a set of task-specific LoRA adapters, our objective is to design a unified deployment framework that (i) enables shared quantization parameters across multiple LoRAs, (ii) supports dynamic runtime injection of LoRA parameters, eliminating the need for multiple task-specific binaries and graphs, and (iii) preserves generative quality under low-bit inference on edge accelerators. This formulation aims to bridge the gap between parameter-efficient adaptation and practical on-device deployment of multi-use-case generative models.

In this work, we propose a method and framework for adapter switching at runtime deployment framework that enables multiple GenAI tasks to share a single compiled model on edge devices. Our key idea is to treat LoRA weights as runtime inputs rather than embedding them into the compiled graph. This design allows task-specific adapters to be dynamically loaded and applied during inference without recompilation. To ensure consistent performance across tasks, we further apply knowledge distillation followed by quantization-aware fine-tuning so that all adapters share compatible quantization parameters.

We implement this using our QUAD framework — a lightweight, unified deployment pipeline that supports fast task switching, low memory footprint, and compatibility with  NPU runtime. Compared to traditional approaches, we observe up to 4× improvement in runtime latency and over 6× reduction in memory usage. Our system also supports over-the-air (OTA) delivery of new LoRAs, enabling future extensibility without changing the base model.


\begin{figure*}
    \centering
    \includegraphics[width=\linewidth]{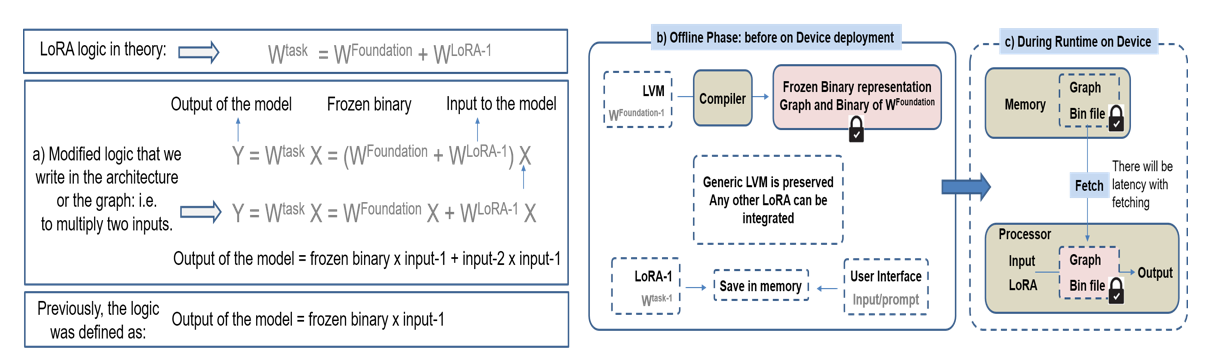}
    \caption{Pictorial representation of proposed approach depicting a) modification of logic to ensure LoRA as input in the architecture, b) offline construction of a single frozen graph that will serve multiple tasks and c) on-device runtime scenario with the proposed approach}
    \label{fig:fig3}
\end{figure*}

\section{Related Work}

\textbf{Large Vision Models and Task Unification.}
Diffusion-based Large Vision Models (LVMs) have significantly advanced visual generation and editing. Latent Diffusion Models (LDMs)~\cite{rombach2022high} enable efficient generation in latent space, while ControlNet~\cite{zhang2023adding} introduces structured conditioning for controllable image editing. Other works such as Palette~\cite{saharia2022palette} explore high-quality image-to-image translation. Recent research attempts to unify multiple tasks within a single generative framework~\cite{kesim2024multi, li2025dual, le2025one, fu2025univg}. However, these approaches typically rely on end-to-end multi-task training and do not support modular runtime adaptation. In practical deployments, particularly on mobile devices, task-specific models are still commonly used, increasing storage and deployment complexity.

\textbf{Adapter-Based Fine-Tuning.}
Parameter-efficient fine-tuning methods such as Low-Rank Adapters (LoRAs)~\cite{hu2022lora} enable task adaptation by learning low-rank updates while keeping the base model frozen. Subsequent works extend adapters to vision models and multi-task settings~\cite{chen2022adaptformer, chen2024conv, wang2025lifelong}. These approaches are widely used in diffusion pipelines to incorporate styles or domain-specific behaviors without retraining the base model. However, in most deployment pipelines LoRA weights are merged into the model prior to inference, resulting in static computation graphs that prevent runtime task switching.

\textbf{Quantization and Edge Deployment.}
Several works explore quantization and efficient inference for large models. QLoRA~\cite{dettmers2023qlora} and QaLoRA~\cite{xu2023qa} improve memory efficiency during fine-tuning through quantization-aware techniques. Edge-focused research such as MobileDiffusion~\cite{zhao2024mobilediffusion} reduces diffusion inference cost for mobile devices, while surveys on edge diffusion deployment~\cite{zheng2025diffusion} highlight the challenges of modular and adaptive inference pipelines. However, existing systems generally assume static model graphs and fixed adapter integration.

\textbf{Our Distinction.}
In contrast to prior work, we enable dynamic LoRA integration for edge GenAI. Our framework treats LoRA weights as runtime inputs to a frozen, quantized LVM, allowing multiple tasks to share a single compiled model without recompilation. By combining knowledge distillation with quantization-aware fine-tuning, our approach supports efficient multi-task inference with runtime-swappable adapters on edge devices.

\section{Proposed Method}

\begin{figure*}
    \centering
    \includegraphics[width=\linewidth]{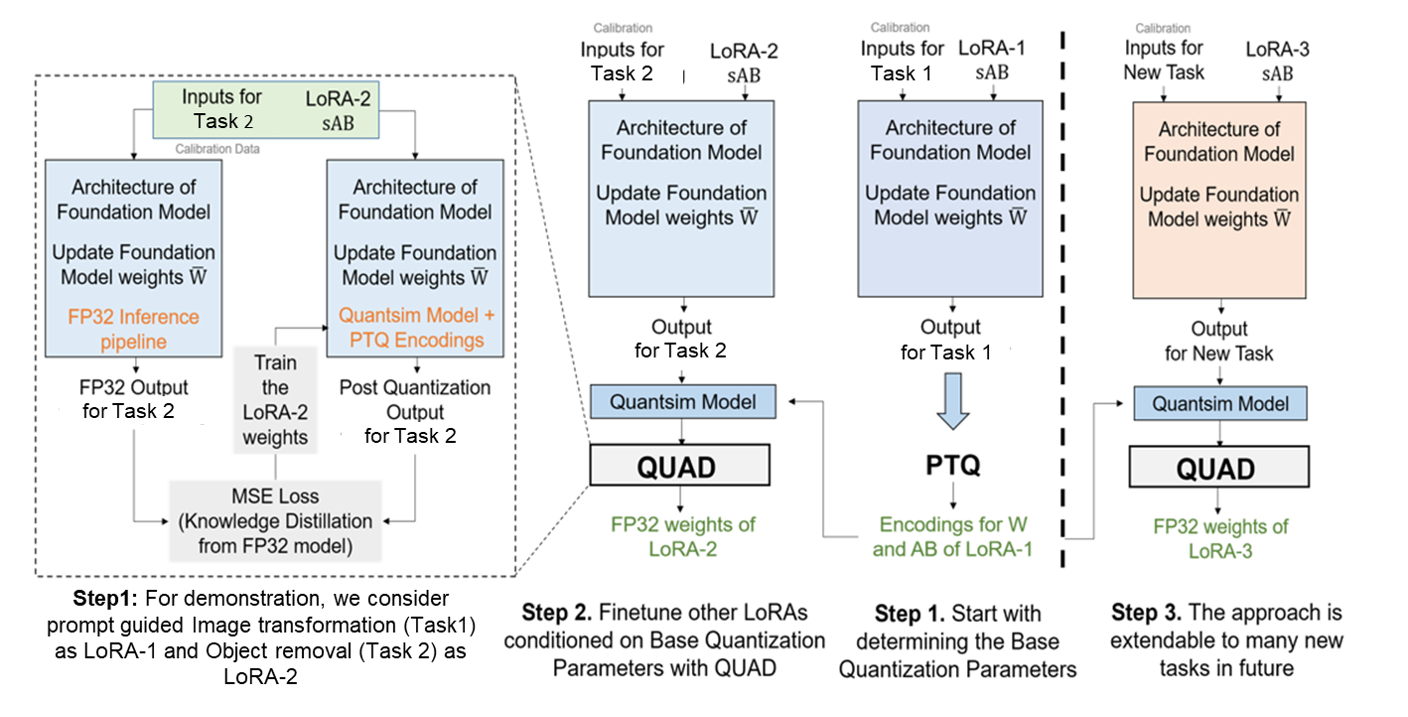}
    \caption{Schematic for a simplified version of compilation and deployment of LVM and depicting the problem definition}
    \label{fig:fig4}
\end{figure*}

\begin{figure}
    \centering
    \includegraphics[width=0.7\linewidth]{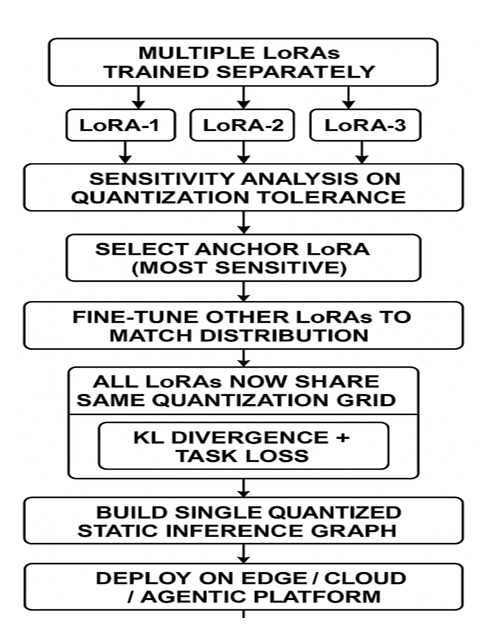}
    \caption{Flowsheet of QUAD framework}
    \label{fig:fig5}
\end{figure}

\begin{figure*}
    \centering
    \includegraphics[width=\linewidth]{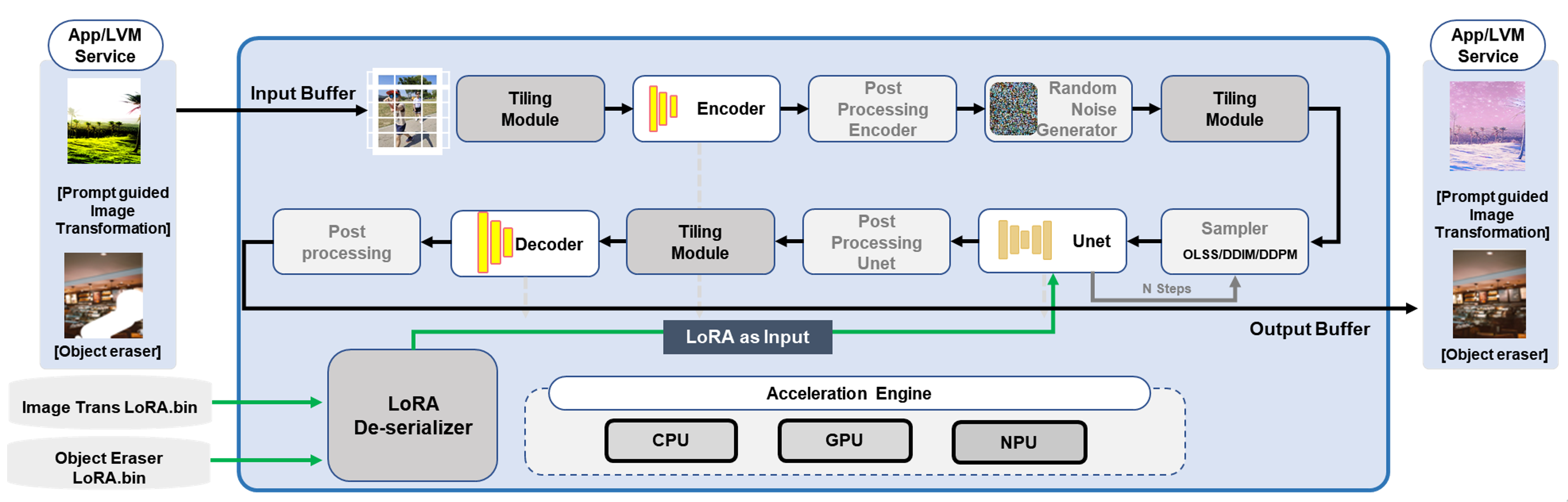}
    \caption{Software stack for facilitating one-for-all LVMs on embedded devices such as smartphones. In this schematic, we show the stack for on-device deployment of two use-cases with a single LVM as described in Section 4}
    \label{fig:fig6}
\end{figure*}
\subsection{LVM and Multi-LoRA Adaptation}

We consider a vision model based on a latent diffusion backbone, similar in structure to Stable Diffusion (SD) 1.5. The model operates in latent space, which reduces the input dimensionality and improves runtime efficiency. It consists of an encoder E, a denoising U-Net backbone U, and a decoder D. The forward pass of LVM can be written as shown in Eq. ~\ref{eq:eq1}, where, x is the input image, $\epsilon_t$ is the sampled noise added during diffusion at time step t, $z_t$ is the latent noisy encoding at time step t, and c represents the conditioning (textual prompt or image).

\begin{equation}
\label{eq:eq1}
\hat{x} =D(U(z_t,c)),z_t=E(x)+ \epsilon_t 
\end{equation}


The U-Net $U$ consists of multiple transformer and convolution blocks where LoRA adapters are introduced. For a given linear transformation $W \in \mathbb{R}^{d_{out} \times d_{in}}$, a LoRA-augmented version is shown in Eq.~\ref{eq:eq2}, where $A \in \mathbb{R}^{d_{out} \times r}$ and $B \in \mathbb{R}^{r \times d_{in}}$, $r \ll \min(d_{out}, d_{in})$ is the rank, and $\alpha$ is the scaling factor.

\begin{equation}
\label{eq:eq2}
W_{LoRA}=W+\alpha AB
\end{equation}

During multi-LoRA training, we freeze the base model parameters and train separate pairs $(A_i,B_i)$ for each task i. At deployment, however, each LoRA typically needs to be integrated into the graph or loaded as a separate variant of the model, as explained in Figure~\ref{fig:fig1}.

\subsection{LoRA as Input and one-for-all Frozen Graph}
In the standard LoRA-based adaptation, each task-specific LoRA is trained separately and inserted into the model statically. This results in multiple variants of the model, each compiled separately for deployment. Such duplication increases ROM usage, complicates updates, and prohibits runtime flexibility. To overcome this, we propose a formulation (see Figure. ~\ref{fig:fig3}), where LoRA weights are treated as runtime inputs, decoupling them from the model’s graph.

For a given linear projection W, we modify the layer to accept LoRA weights as input tensors A and B. The computation transforms to Eq. ~\ref{eq:eq3}, where, W is frozen weight binary, x and y are input and output maps, respectively. 

\begin{equation}
\label{eq:eq3}
y=Wx+\alpha A(Bx) 
\end{equation}

As shown in ~\ref{fig:fig3}a, instead of compiling separate binaries, we treat LoRA weights as input tensors, modifying the LVM architecture to support dynamic LoRA switching at runtime using a frozen graph (~\ref{fig:fig3}b). During inference, user inputs trigger LoRA weight loading from memory, enabling task-specific outputs (~\ref{fig:fig3}c). This modification requires changing the original graph such that each LoRA-augmented layer exposes additional input nodes for A and B. The model compilation is performed only once, and at inference time, different tasks are supported by supplying corresponding LoRA weights on-the-fly. 
This formulation introduces the following advantages:
\begin{itemize}
    \item Single compiled binary: Only one foundation model graph needs to be compiled and deployed.
    \item Dynamic task switching: Different LoRAs can be loaded at runtime without reloading LVM from memory.
    \item LoRA modularity: LoRA weights can be updated or added in future without affecting the base model binary.
\end{itemize}

In terms of runtime implementation, LoRA weights are streamed as buffers and bound to the model inputs using lightweight APIs during inference initialization. Since the model is pre-frozen and graph-optimized, inference can proceed with minimal overhead. This reformulation enables the subsequent quantization and deployment stage, making the system suitable for edge devices.

\subsection{QUAD: Unified LoRA-Compatible Deployment}
After restructuring the model to accept LoRA weights as runtime inputs, the next challenge is deploying this flexible model efficiently on hardware-limited devices. Standard quantization pipelines for DSP architectures such as the NPUs, assume a fixed set of quantization parameters. Introducing dynamically injected LoRAs breaks this assumption — especially if each LoRA has its own distribution and scaling resulting in different quantization parameters. To solve this, we propose the QUAD (Quantization with Unified Adaptive Distillation) framework. The goal is to enable a single quantization profile that is shared across all LoRAs and the base model, ensuring compatibility and runtime efficiency.

\textbf{Quantization Setup}. For a given tensor T, the scale s and zero-point z are defined as shown in Eq. ~\ref{eq:eq4} and ~\ref{eq:eq5}, respectively, where the $T_{max}$ and $T_{min}$  is obtained from the Tensor, and $q_{max}$ and $q_{min}$ are bounds of quantization (e.g., for signed INT-8 quantization, $q_{max}= 127$ and $q_{min}= -128$). The quantized representation $\hat{T}$ is defined as shown in Eq. ~\ref{eq:eq6}.  

\begin{equation}
\label{eq:eq4}
s=(T_{max}-T_{min})/(q_{max}-q_{min} ) 	
\end{equation}

\begin{equation}
\label{eq:eq5}
z=\lfloor(T_{min}/s)-q_{min} \rfloor
\end{equation}

\begin{equation}
\label{eq:eq6}
\hat{T} =clip(\lfloor(T-z)/s\rfloor,q_{min},q_{max} )  
\end{equation}

\begin{equation}
\label{eq:eq7}
T \sim s.\hat{T} +z
\end{equation}

In QUAD, we enforce shared scale s and shared zero-point z across all LoRA weights $A_i, B_i$ and the base weight matrix W. This required aligning the distributions of all LoRAs at training time. However, since the LoRAs are built independent of each other at training time, it is our responsibility to finetune the trained LoRAs such that their distributions result in shared quantization parameters at inference time. 

\textbf{Unified Quantization Strategy}. First, the base quantization parameters (scale and offset) have to be determined. We perform sensitivity analysis across all LoRA adapters to identify the most “quantization-sensitive” LoRA (the LoRA whose accuracy drop is maximum when perturbed for fine-tuning), which then serves as the anchor for defining fixed quantization parameters.

Let $f(x; w)$ and $f(x;\Tilde{w})$ be the output of LoRA and quantized LoRA, respectively for a given input x. Then we define the Quantization Sensitivity Score (QSS) for the LoRA as shown in Eq. ~\ref{eq:eq8}, where $E_x$ is expectation over the data x, and $D(.||.)$ is a suitable Divergence metric (e.g., Jenson Shanon). Lower QSS implies less sensitive to Quantization. The quantization parameters of the LoRA with highest QSS are then taken as the fixed quantization parameters

\begin{equation}
\label{eq:eq8}
QSS=E_x [D(f(x; w)||f(x; \Tilde{w}))] 
\end{equation}

In a case when all LoRAs are equally sensitive, we propose the fallback (Unified-LoRA) to determine the quantization parameters. Instead of selecting the sensitive LoRA, we compute global quantization parameters based on the merged weight distributions of all LoRA adapters ensuring an unbiased approach towards all LoRA adapters.



\textbf{Knowledge Distillation based Finetuning.}
After determining the shared quantization parameters, the LoRAs are finetuned to operate under this unified quantization profile through knowledge distillation, as illustrated in Figure~\ref{fig:fig4}. Consider the case where the quantization parameters obtained from LoRA-1 (the LoRA with the highest QSS) must be enforced on LoRA-2. To achieve this, we construct a quantization simulation (QuantSim) model for the LVM with LoRA-2, where the weights are quantized using the PTQ encodings derived from LoRA-1.

The QuantSim model produces the output of the quantized LVM + LoRA-2, which is compared with the output of the corresponding full-precision model. The full-precision network acts as the teacher, while the quantized model serves as the student. The LoRA parameters are optimized by minimizing a reconstruction loss between the teacher and student outputs, combined with the original LVM training objective.

Through iterative optimization, the LoRA weights are adapted to satisfy the shared quantization parameters while preserving task performance. Since this process aligns independently trained LoRAs under a unified quantization configuration via distillation, we refer to it as \textit{Quantization with Unified Adaptive Distillation (QUAD)}.

The proposed approach is generic and can be applied to any AI model as presented in the flowsheet in Figure ~\ref{fig:fig5}. The QUAD framework thus enables:
\begin{itemize}
    \item Compatibility across LoRAs under one quantized graph.
    \item Reduced memory by avoiding multiple graphs.
    \item Fast runtime switching without reloading the binaries.
\end{itemize}
This completes the model-level preparation. The next step is deployment, which includes graph optimization, conversion, and runtime orchestration.

\subsection{Compilation and Deployment: Graph Optimization}

After quantization and restructuring the model to accept LoRA weights at runtime, the final stage involves compiling the model into a hardware-executable format and deploying it on edge devices. This stage focuses on optimizing the computation graph, memory layout, and operator compatibility for efficient inference.

\textbf{Model Conversion.} 
The training-time model (typically implemented in PyTorch) is first exported to an intermediate representation (IR) using ONNX and subsequently converted into a hardware-specific IR through vendor toolchains. During this process:
\begin{enumerate}
    \item Inherent parallelism is introduced (e.g., linear layers mapped to convolution and multi-head attention decomposed into independent heads).
    \item Static operator fusion is applied (e.g., convolution + activation).
    \item Placeholder inputs are added for LoRA weight tensors.
    \item Quantization nodes are inserted using calibration parameters obtained from QUAD.
\end{enumerate}

\textbf{Graph Optimizations.} 
The converted graph undergoes additional transformations to improve runtime efficiency:
\begin{itemize}
    \item \textit{Scale folding:} quantization scale and zero-point parameters are merged into adjacent layers to reduce arithmetic operations.
    \item \textit{Constant folding:} precomputed constants are embedded into the graph to avoid runtime computation.
    \item \textit{Dead code elimination:} unused branches (e.g., fallback LoRA paths) are removed.
\end{itemize}

\textbf{Memory Layout and LoRA Binding.} 
During inference, the runtime loads the frozen base model and dynamically injects LoRA weights into designated input slots. The LoRA tensors are stored as buffers in RAM and bound to the model inputs through lightweight APIs, enabling task-specific inference without graph recompilation.

\textbf{Runtime and Software Stack.} 
The compiled model executes within a lightweight runtime environment (Figure~\ref{fig:fig6}), which can be integrated into standard mobile AI stacks. The runtime consists of three main components:
\begin{itemize}
    \item \textbf{Graph runtime:} executes the quantized model IR.
    \item \textbf{LoRA loader:} manages LoRA weight loading and buffer binding.
    \item \textbf{Scheduler:} optimizes thread utilization and memory reuse across inference calls.
\end{itemize}

This modular design supports extensions such as LoRA caching, background preloading, and real-time task switching across applications, while remaining portable across hardware platforms that support quantized operators.

\section{Experiments and Results}


We evaluate the proposed approach through both accuracy and on-device performance analysis. Specifically, we compare the outputs of the full-precision (FP32) model executed on an x86 server with those of the quantized model deployed on-device. In addition to accuracy metrics, we report key on-device performance indicators, including latency, memory usage, and runtime accuracy, measured across multiple chipsets (Qualcomm, MediaTek, and LSI). These experiments provide a comprehensive evaluation of the effectiveness and deployment efficiency of QUAD framework.

\begin{table}[]
\caption{Accuracy analysis of quantized model. The prompt guided image transformation use-case is considered as base LoRA and the effect of quantization is studied in its case. In case of Object removal, QUAD is applied and thus the effect of enforcing the shared quantization parameters is presented.}
\label{tab:tab1}
\resizebox{0.5\textwidth}{!}{
\begin{tabular}{lll}
\toprule
\multicolumn{1}{l}{\textbf{Usecase}}  & {\textbf{Metric}} & \textbf{FP32 vs INT8} \\
\midrule
\textbf{Prompt   }   & $sim_{d}$                     & 0.9428                                         \\
\textbf{guided}                                                            & $sim_{image}$                & 0.881                                          \\
        \textbf{Image}                                                      & Structure   loss           & 0.045                                          \\
        \textbf{Transformation}                                                       & Custom   clip score        & 0.008                                          \\
\midrule
\textbf{Object}   & FID                        & 5.5287                                         \\
\textbf{Removal}                                                               & LPIP                       & 0.12                                           \\
        \textbf{usecase}                                                       & SSIM                       & 0.94                                          \\
                                                               & PSNR                       & 33.04     \\
\bottomrule
\end{tabular}
}
\end{table}

\begin{figure}
    \centering
    \includegraphics[width=0.8\linewidth]{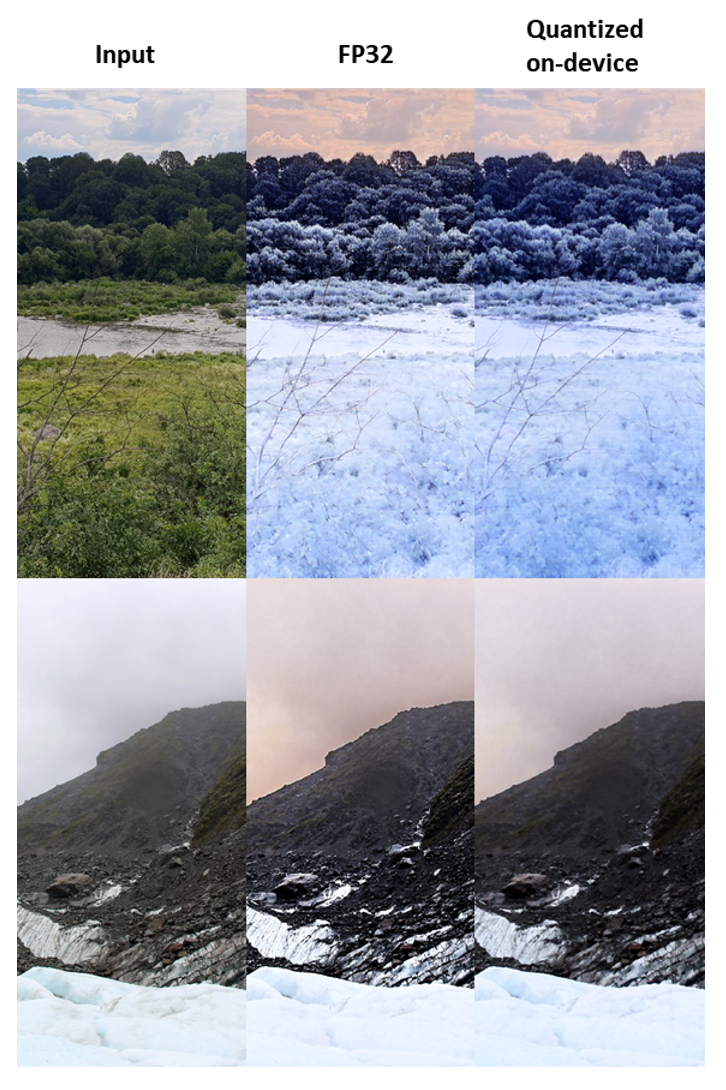}
    \caption{Accuracy comparison in terms of Image Quality between the FP32 model on server and Quantized model deployed on device for the prompt guided Image Transformation use-case}
    \label{fig:fig7}
\end{figure}

\begin{figure*}
    \centering
    \includegraphics[width=\linewidth]{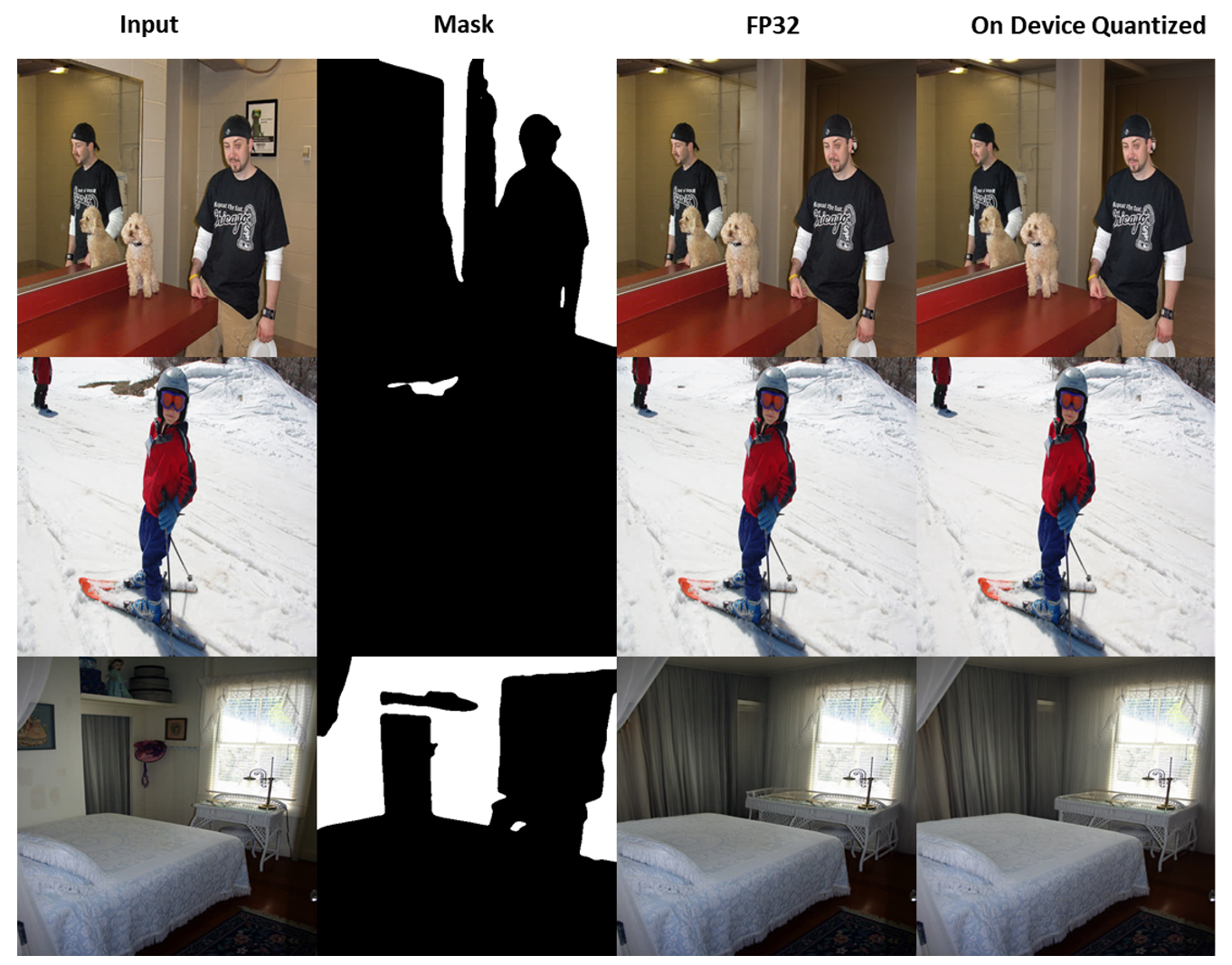}
    \caption{Accuracy comparison in terms of Image Quality between the FP32 model on server(x86) and Quantized model deployed on device (GS25) for the object removal use-case}
    \label{fig:fig8}
\end{figure*}

Given the trained LVM and LoRAs for different use-cases, in the first set of experiments, it was necessary to determine the same quantization parameters for all LoRAs. Prompt guided Image Transformation use-case was considered as the primary based on the QSS score and the quantization encodings of Object removal LoRA were fine-tuned using the proposed QUAD strategy. The comparison between the full precision model’s output on server (x86) and the quantized model’s output on-device resulted in an FID score of 5.5287. The quantization accuracies were also compared in case of prompt guided Image Transformation use-case with the full precision model and presented in Table ~\ref{tab:tab1}, where $sim_d$ is s metric based on cosine similarity between direction vectors of quantized and full-precision latent representations, and $sim_{image}$ measures the semantic similarities between the corresponding latent representations. The results of visual comparison between the x86 full precision and on-device quantized outputs for prompt guided Image Transformation and Object removal are presented in Figure ~\ref{fig:fig7} and ~\ref{fig:fig8}, respectively. 

\begin{table*}[]
\caption{On-Device Key Performance Indicators (memory and latency) of 2 use-cases with our proposed QUAD framework and inference stack with multi-LoRA as input and a SD 1.5 based 1.1B model as Foundation Model.}
\label{tab:tab2}
\resizebox{\textwidth}{!}{
\begin{tabular}{lp{2.5cm}lp{2.5cm}lp{2.5cm}l}
\toprule
\textbf{Chipset}                  & \multicolumn{2}{c}{\textbf{Qualcomm (GS25)}} & \multicolumn{2}{c}{\textbf{LSI (GS25)}} & \multicolumn{2}{c}{\textbf{Mediatek (Tab S11)}} \\
\midrule
\textbf{Scenario}                 & \textbf{Prompt Guided Image Transformation}       & \textbf{Object Removal}      & \textbf{Prompt Guided Image Transformation}    & \textbf{Object Removal}    & \textbf{Prompt Guided Image Transformation}        & \textbf{Object Removal}        \\
\midrule
\textbf{Sampler}                  & OLSS             & LCM       & OLSS         & ED      & OLSS              & ED          \\
\textbf{Steps}                  & 8              & 4       & 8         & 4      & 8             & 4          \\
\textbf{VAE Encoder execute (ms)} & 449                      & 458               & 427                   & 425             & 639                       & 643                 \\
\textbf{Unet execute (ms)}        & 250                      & 249               & 409                   & 175             & 509                       & 254                 \\
\textbf{VAE Decoder execute (ms)} & 839                      & 836               & 896                   & 898             & 1090                      & 1082                \\
\textbf{End-to-End (ms)}          & 8826                     & 3723              & 12456                 & 4217            & 15682                     & 5528                \\
\textbf{Shared FM ROM (MB)}       & \multicolumn{2}{c}{1375}                     & \multicolumn{2}{c}{1125}                & \multicolumn{2}{c}{1177}                        \\
\textbf{LoRA ROM (MB)}            & 119                      & 119               & 134                   & 104             & 31                        & 87                  \\
\textbf{Peak   RAM (MB)}          & 1739                     & 1873              & 1259                  & 1229            & 1590                      & 1494               \\
\bottomrule
\end{tabular}
}
\end{table*}

\begin{table}[]
\caption{On-Device Key Performance Indicators (memory and latency) of 4 use-cases with our proposed QUAD framework and inference stack with multi-LoRA as input and a SD 1.5 based 0.7B parameter model as Foundation model(OLSS sampler-8 steps)}
\label{tab:tab3}
\resizebox{0.45\textwidth}{!}{
\begin{tabular}{lp{1cm}p{1cm}p{1cm}p{1cm}}
\toprule
\textbf{Scenario}                 & \textbf{Text-to-Image} & \textbf{Sketch-to-Image} & \textbf{Sticker Generation} & \textbf{Portrait Studio} \\
\midrule
\textbf{VAE Encoder execute (ms)} & NA                     & 68                       & NA                          & NA                       \\
\textbf{Unet execute (ms)}        & 48                     & 48                       & 48                          & 51                       \\
\textbf{VAE Decoder execute (ms)} & 150                    & 152                      & 152                         & 149                      \\
\textbf{End-to-End (ms)}          & 1052                   & 1527                     & 1080                        & 1874                     \\
\textbf{Shared FM ROM (MB)}       & \multicolumn{4}{c}{844}                                                                                    \\
\textbf{LoRA ROM (MB)}            & NA                     & 77                       & 77                          & 77                       \\
\textbf{Peak   RAM (MB)}          & 2024                   & 2348                     & 2019                        & 2176                    \\
\\
\bottomrule
\end{tabular}
}
\end{table}


Subsequently, the model with LoRAs was converted, compiled and deployed on the NPU of Galaxy S25 and Tab S11 with the proposed novel inference stack (see Figure ~\ref{fig:fig6}) that enabled dynamic runtime switching between the LoRAs. The on-device performance metrics are presented in Table ~\ref{tab:tab2} and ~\ref{tab:tab3}. It is important to note that the proposed method works for any hardware (Qualcomm / Exynos / Mediatek), thus making it chipset-agnostic.  
The proposed approach of a) LoRA as input, b) QUAD framework and c) inference stack for deployment of multi-LoRA based one-for-all foundational model was also applied for four novel use-cases. The on-device KPIs are presented in Table ~\ref{tab:tab3}.
With the proposed approach, we get the maximum memory benefit when the number of use-cases increases significantly in the future. For example, a latent diffusion model of size 1.4GB combined with ten LoRA modules (~120MB each) would require ~15GB when compiled separately. In contrast, our method requires only 2.6GB by sharing a single base model with runtime LoRA loading, resulting in a 6× reduction in memory footprint. Even in current deployments with just 2 use-cases, we observe performance gains. When separate graphs are fetched and swapped from memory for each use-case (, the overhead becomes significant. By maintaining a single model graph and switching only LoRA weights at runtime, our system saves ~1.5 seconds in end-to-end latency during inference compared to the traditional per-graph approach.

\section{Ablation Analysis}

\begin{table}[]
\caption{Ablation analysis. Comparison of accuracy measured on-device (GS 25) in prompt guided Image Transformation use-case when quantized with INT8 precision and mixed bit precision. Proposed model is quantized with W8A16. Each scenario x:y implies x\% params in W8A8 and y\% params in in W8A16}
\label{tab:tab4}
\resizebox{0.45\textwidth}{!}{%
\footnotesize
\begin{tabular}{lllll}
\toprule
\textbf{Scenario}                     & \textbf{FID} & \textbf{LPIPS} & \textbf{PSNR} & \textbf{SSIM} \\
\midrule
\textbf{0:100}                & 12.2308      & 0.1083         & 32.712        & 0.9808        \\
\textbf{20:80} & 12.45        & 0.1086         & 32.6766       & 0.9806        \\
\textbf{40:60} & 13.0494      & 0.1086         & 32.6819       & 0.9806        \\
\textbf{60:40} & 12.7991      & 0.109          & 32.5978       & 0.9803        \\
\textbf{80:20} & 14.2763      & 0.1125         & 31.4065       & 0.9777        \\
\textbf{90:10} & 26.6183      & 0.1372         & 28.8822       & 0.9628        \\
\textbf{100:0}                & 599.0724     & 0.699          & 5.4424        & 0.2324        \\
\bottomrule
\end{tabular}
}
\end{table}

\begin{table}[]
\caption{Ablation analysis. Comparison of accuracy and on-device KPIs measured on Tab S11 in Object removal use-case when quantized with INT8 precision and mixed bit precision. Proposed model is quantized with W8A16. Each scenario x:y implies x\% params in W8A8 and y\% params in in W8A16.}
\label{tab:tab5}
\resizebox{0.5\textwidth}{!}{%
\begin{tabular}{p{1cm}p{1cm}p{1cm}p{1cm}p{1cm}p{1cm}p{1cm}}
\toprule
\textbf{Scenario}                & \textbf{Init Time (ms)} & \textbf{Execute Time (ms)} & \textbf{LoRA ROM size (MB)} & \textbf{FID} & \textbf{SSIM} & \textbf{PSNR} \\
\midrule
\textbf{0:100}                    & 1156                    & 469                        & 181                         & 25.884       & 0.8064        & 25.28         \\
\textbf{10:90}   & 1144                    & 450                        & 158                         & 27.419       & 0.7951        & 25.32         \\
\textbf{20:80}   & 1101                    & 438                        & 138                         & 27.106       & 0.7951        & 25.23         \\
\textbf{30:70} & 1097                    & 405                        & 138                         & 27.108       & 0.7941        & 25.18        \\
\bottomrule
\end{tabular}
}
\end{table}

\textbf{Quantization accuracy.} The LVM model and corresponding LoRAs were quantized with model weights in INT8, LoRA weights in INT16 and activations in INT16 precision (W8A16). However, in quest of further reduction in memory and latency, we conducted an exhaustive study to reduce the bit precision of activations to INT8 (W8A8) specifically in prompt guided Image Transformation use-case. Further, we also tried mixed precision quantization with a fixed \% of U-Net’s activation layers in INT8 precision. The results of this study are presented in Table ~\ref{tab:tab4} and conclude that the W8A16 yields the best result. We have also experimented with LoRA weights in INT8 precision reducing the ROM by 1.5 folds with marginal drop in accuracy.   

\textbf{On-device KPI.} Another analysis was conducted to study the impact of mixed-precision quantization on the on-device KPIs when profiled on Tab S11 for the Object removal use-case. The corresponding measures for accuracy, latency and memory are presented in Table ~\ref{tab:tab5}, indicating a clear trade-off between accuracy and on-device KPIs.

\section{Conclusion}

We presented a unified deployment framework for supporting multiple GenAI tasks on edge devices using a single visual foundation model. By treating LoRA weights as runtime inputs, the system enables dynamic task switching without recompiling or duplicating the model graph. Combined with the QUAD framework for shared quantization across LoRAs, the approach enables efficient multi-task inference on resource-constrained hardware. Experiments across multiple chipsets demonstrate up to $6\times$ memory reduction and improved latency while maintaining high output quality.

{
    \small
    \bibliographystyle{ieeenat_fullname}
    \bibliography{main}
}


\end{document}